\documentclass[twoside,twocolumn]{article}

\usepackage[round]{natbib} 
\renewcommand{\cite}[1]{\citep{#1}}

\usepackage{apptools}

\usepackage{amsmath}
\usepackage{amsfonts}
\usepackage{amssymb}
\usepackage{amsthm}

\usepackage{algorithm}
\usepackage{algpseudocode}
\usepackage{hyperref}
\usepackage{cleveref}

\usepackage{graphicx}
\usepackage{wrapfig}
\usepackage{float}
\usepackage{subcaption}

\usepackage[all]{xy}
\usepackage{tikz}
\usepackage{pifont}

\theoremstyle{definition}
\newtheorem{definition}{Definition}

\theoremstyle{plain}
\newtheorem{lemma}{Lemma}
\newtheorem{theorem}{Theorem}
\newtheorem{corollary}{Corollary}

\theoremstyle{remark}
\newtheorem{remark}{Remark}

\usepackage{authblk}

\title{A probability theoretic approach to\\ drifting data in continuous time domains}
\author{Fabian Hinder}\author{Andr\'e Artelt}\author{Barbara Hammer}
\affil{Cognitive Interaction Technology (CITEC) \\
	Bielefeld University \\
	Inspiration 1, D-33619 Bielefeld, Germany\\
	\texttt{\{fhinder,aartelt,bhammer\}@techfak.uni-bielefeld.de}}
\begin{document}
	\maketitle

	\begin{abstract}%
		%
		The notion of drift refers to
		the phenomenon that the distribution, which is underlying the observed data,
		changes over time.
		Albeit many attempts were made to deal with drift, formal notions of
		drift are application-dependent
		and  formulated in various degrees
		of abstraction and mathematical coherence.
		In this contribution, 
		we provide a probability theoretical framework,
		that allows a formalization of drift in continuous time,
		which subsumes popular notions of drift.
		In particular, it sheds some light on common practice such as change-point detection or machine learning methodologies in the presence of drift. 
		It
		gives rise to a new characterization of drift in terms
		of stochastic dependency between data and
		time. This particularly intuitive formalization enables
		us to design a new,  efficient drift detection method. Further, it  induces a technology, to decompose observed 
		data into a drifting and a non-drifting part.
		
	\end{abstract}
	
	\section{INTRODUCTION}
	\label{sec:intro}
	One  fundamental assumption in classical machine learning is the fact that observed data are i.i.d.\ according to some unknown underlying probability measure $P_X$, i.e.\ the data generating process is stationary. 
	Yet, this assumption is often violated 
	as soon as machine learning faces real world problems: models are subject to seasonal changes, changed demands of individual costumers, ageing of sensors, etc. In such settings, 
	life-long model adaptation rather than classical batch learning is required for optimum performance.
	Since drift, i.e.\ the fact that data is no longer identically distributed, is a major issue in many real-world
	applications of machine learning,
	many attempts were made to deal with this setting \cite{DBLP:journals/cim/DitzlerRAP15}.
	
	Depending on the domain of data and application, the presence of drift is 
	modelled in different ways.
	As an example, covariate shift  refers to the  situation of  training and test set having different marginal distributions   \cite{5376}. Learning for data streams extends this setting to an unlimited (but usually  countable) stream of observed data, mostly in supervised learning scenarios \cite{asurveyonconceptdriftadaption}.
	Here one distinguishes virtual and real drift, i.e.\ non-stationarity of the marginal distribution only or also the posterior. Learning technologies for such situations often rely on windowing techniques, and adapt the model based on the characteristics of the data in an observed time window. Active methods explicitly detect drift, usually referring to drift of the classification error, and trigger model adaptation this way, while passive methods continuously adjust the model \cite{DBLP:journals/cim/DitzlerRAP15}. 
	
	Data streams also occur naturally whenever times series are dealt with, such as time series prediction. Unlike streaming  data  as considered by \citet{DBLP:journals/cim/DitzlerRAP15} or \citet{asurveyonconceptdriftadaption}, 
	time 
	series modeling relies on the assumption of a direct functional relation of subsequent observations.  One  distinguishes stationary and non-stationary time series, and one particularly interesting challenge  is change point detection, i.e.\ time points where abrupt variations are observed
	\cite{Aminikhanghahi:2017:SMT:3086013.3086037,DBLP:journals/tnn/AlippiBR17}
	
	Interestingly, the overwhelming majority of such drift learning approaches deals with 
	discrete time processes rather than continuous time \cite{DBLP:journals/tnn/Roveri19}. Further, the majority refers to supervised learning scenarios with an emphasis on minimization of a cost function such as the interleaved train-test error. Only first approaches 
	consider the particularly relevant question how to substantiate such models by
	methods for  understanding drift \cite{DBLP:journals/corr/WebbLPG17}.
	
	The purpose of this contribution is to provide a proper probabilistic definition of drift for streaming data in continuous time, which subsumes common definitions of drift in the literature. Unlike \citet{DBLP:journals/kais/GoldenbergW19}, we are not interested how to identify and measure different types of drift (such as real drift, virtual drift, reoccurring drift, etc.); rather, we are interested in a unifying probabilistic model of drift in continuous time processes, which also justifies common practice to deal with drift, such as identifying change points or learning from time windows. 
	
	Now, we will introduce a measure-theoretic setting to define drift in continuous time first, and we  introduce different notions of drift from
	the literature and show their equivalence. Then, we establish a new characterization by relating drift to an independence criterion of time and data, giving rise to particularly efficient drift detection models as well as an elegant way to disentangle drifting and non-drifting parts in observed data. We demonstrate these methods  in experiments.
	
	\section{A THEORY OF DRIFT}
	\label{theory}
	
	In the following we will define the notion of a drift process. Afterwards we will give several definitions of drift, that have been proposed in different fields, and investigate their relationships. In particular we introduce a new definition of drift for continuous time processes in  Section~\ref{SecDriftAsDependency}.
	Due to space restrictions, all proofs (and some explanations of well known definitions) are contained in the auxiliary material (identifiable by numeration starting with an "A").
	
	\subsection{Definition of a drift process}
	
	\newcommand{\X}{\mathfrak{X}}
	\newcommand{\T}{\mathfrak{T}}
	\newcommand{\A}{\mathcal{A}}
	\newcommand{\B}{\mathcal{B}}
	\newcommand{\Prob}{\mathbf{P}}
	\renewcommand{\d}{\textnormal{d}}
	\renewcommand{\P}{\mathbb{P}}
	
	\newcommand{\Q}{\mathbb{Q}}
	\newcommand{\R}{\mathbb{R}}
	\newcommand{\N}{\mathbb{N}}
	\newcommand{\Borel}{\mathfrak{B}}
	\newcommand{\E}{\mathbb{E}}
	
	\newcommand{\argmin}{\textnormal{argmin}}
	\newcommand{\argmax}{\textnormal{argmax}}
	
	\newcommand{\textdef}[1]{\textit{#1}}
	
	In the usual, time invariant setup of machine learning one considers a generative process $P_X$, i.e. a probability measure, on a measurable space $(\X,\A)$. In this context one views the realizations of a $\X$-valued, $P_X$ distributed random variable $X$ as samples.
	Depending on the objective, learning algorithms try to infer the data distribution based on these samples or, in the supervised setting, a  posterior distribution. We will not distinguish these settings and only consider  distributions in general, this way subsuming the notion of both, real drift and virtual drift.
	
	Many processes in real-world applications are not time independent, so it is reasonable to incorporate time into our considerations. One prominent way to do so is to consider an index set $\T$, representing time, and a collection of probability measures $p_t$ on $\X$ indexed over $\T$ \cite{asurveyonconceptdriftadaption}. 
	
	In the following we investigate the relationship of those $p_t$, with drift referring to a property of the relationship of several $p_t$ at different time points $t$. A first, and mathematically equivalent, step to do so is by considering $p:\T \to \Prob(\X)$, $t \mapsto p_t$ as a map rather than a conglomerate, here $\Prob(\X)$ denotes the set of all probability measures on $\X$. We will sometimes refer to this as a \textdef{non-probabilistic drift process}.
	
	For continuous time, we need more structure; hence we view $p_t$ in a measure theoretic setup, which yields:
	
	\begin{definition}
		\label{def:DirftProcess}
		Let $(\T,\B)$ and $(\X,\A)$ be two measurable spaces.
		A \textdef{drift process} $(p_t,P_T)$ is a Markov kernel\footnote{See Definition~\ref{def:MarkovKernel}} $p_t$ from $\T$ to $\X$  and a probability measure $P_T$ on $\T$.  
	\end{definition}
	
	When $(\X,\A)$ and $(\T,\B)$ or $P_T$ are clear, we sometimes just write $(P_T,p_t)$ resp. $p_t$ for simplicity.
	Notice that this is a very minor restriction regarding $p_t$ as compared to a non-probabilistic drift process, since we basically only state that we want $t \mapsto p_t(D)$ to be a measurable map for all $D \in \A$\;\footnote{See Remark~\ref{thm:MarkovIsNatural} for more details}.
	
	%
	%
	Note that this notion of drift processes is actually extremely natural:
	\begin{remark}
		\label{FubiniRemark}
		By Fubini's theorem every drift process $(p_t,P_T)$ induces a probability measure $p_t \otimes P_T$\;\footnote{See Remark~\ref{thm:FubiniForMarkovKernel} for definition and more details} on $\X \times \T$.\\
		Conversely under some mild assumptions (e.g.\ $\T$ and $\X$ are polish spaces \cite{Probabilitymeasuresonmetricspaces}), every probability measure $P$ on $\X \times \T$ gives rise to a drift process, i.e. we may find a Markov kernel $p_t$ with $p_t \otimes P_T = P$, where $P_T$ is the marginalization of $P$ onto $\T$. \\
		In particular if $P_T$ has no null sets, i.e.\ $P_T(\{t\}) > 0$ for all $t \in \T$, then we have $p_t = P(\cdot \mid \X \times \{t\})$ the conditional expectation given $t \in \T$.
	\end{remark}
	
	We will now define drift: A very common notion specifies drift
	as the fact that distributions vary over time \cite{asurveyonconceptdriftadaption}, i.e.\ there exist $t,s \in \T$ such that $p_t \neq p_s$. Conversely a process has no drift iff $p_t = p_s$  for all $t,s \in \T$. In the following definition, this notion is adapted to the measure theoretic setup.
	\begin{definition}\label{def2}
		Let $(p_t,P_T)$ be a drift process. We say that $p_t$ has \textdef{no drift} or does not drift if $p_t = p_s$ holds $P_T$-a.s., i.e. $(P_T \times P_T)(\{ (s,t) \in \T\times\T \mid p_t = p_s \}) = 1$\footnote{See Definition~\ref{def:ProductMeasure} for details}. We say that $p_t$ has \textdef{drift} or is drifting if it is not the case that it does not drift.
	\end{definition}
	
	Here we allow differences of distributions in null sets, i.e.\ we allow that $p_t \neq p_s$ if we do not expect to ever observe a sample at $t$ resp.\ $s$, which makes this difference irrelevant for applications. In particular if there exists a measure $P_T$ on $\T$ that has no null set then both notions coincide (see Lemma~\ref{ClassicalAndProbabilisticDrift}).
	
	
	Though our definition is already weaker than the one given in \citet{asurveyonconceptdriftadaption}, it is still too strict for applications. This is mainly caused by the fact that the decomposition described in Remark~\ref{FubiniRemark} is not unique. Therefore we need a notion of drift where we no longer distinguish drift processes that do not differ in this sense. This leads to the following definition:
	
	\begin{definition}
		Let $(p_t,P_T)$ be a drift process. We say that $p_t$ has \textdef{no proper drift} iff there exists a drift process $(p_t',P_T')$, such that ${p_t \otimes P_T = p_t' \otimes P_T'}$, that does not drift. We say that $p_t$ has \textdef{proper drift} iff it is not the case that it has no proper drift.
	\end{definition}
	
	\begin{remark}
		\label{ProperAndNonProperDriftRemark}
		Proper drift implies drift but the converse does not hold in general. However under some assumptions, e.g. $\T$ is (at most) countable or $\X = \R^d$ (or more general if $\A$ is generated by a countable set, stable under finite intersections\footnote{See Definition~\ref{def:GeneratedSigmaAlgebra} for details}
		; see Lemma~\ref{thm:UniquenessOfKernel}), drift and proper drift are equivalent. 
	\end{remark}
	
	
	\subsection{Road map}
\begin{figure}[!htb]
		\centering
		~\vspace{-1em}
		\begin{align*}
		\xymatrix{
			\text{non-stationary SP} \ar@/_/@{=>}[d]_{\text{(1)}} & \\
			\text{drift} \ar@/_/@{=>}[u]_{\text{Theorem~\ref{Stationary}}} \ar@/_/@{=>}[d]_{\text{(2)}} \ar@{<=>}[r]^{\text{\!\!\!\!\!\!\!\!\!\!\!\!\!\!\!\!\!\!\!\!\!\!\!\! Theorem~\ref{ProbabilisticDriftAndConstantDriftAreEquivalent} }} & \text{change of distribution} \\
			\text{proper drift} \ar@/_/@{=>}[u] \ar@{<=>}[r]^{\text{\!\!\!\!\!\! Theorem~\ref{DriftAsDependency} }} \ar@{<=>}[d]^{\text{  Theorem~\ref{ModelDriftAndProperDriftAreEquivalent} }}  & \text{dependency drift} \\
			\text{model drift} \ar@{<=>}[r] & \text{alternating sets} \ar@/_/@{=>}[d]_{\text{ Corollary~\ref{Changepoint}}} \\
			& \underset{(\T = \R)}{\text{change points}} \ar@/_/@{=>}[u]
		}
		\end{align*}
		\caption{Equivalent Notions Of Drift}
		\label{fig:roadmap}
	\end{figure}
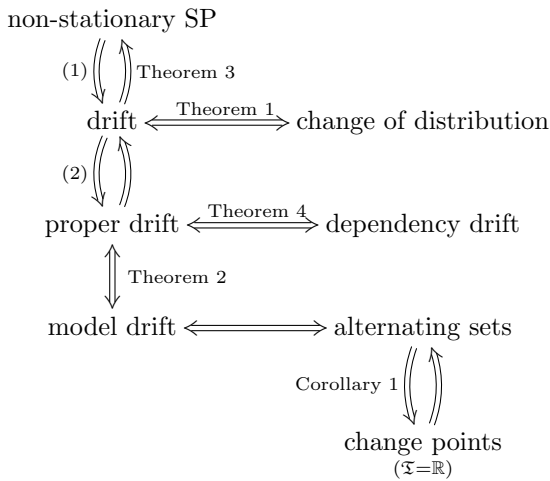

	The results we are going to show, i.e.\ the fact that this notion subsumes several popular definitions from the literature, are summarized in Figure~\ref{fig:roadmap}. %
	
	(1) and (2) do not hold in general. (1) holds for example if $X_t$ has $\P$-a.s. continuous paths.\footnote{See Definition~\ref{def:StochProcess}} (2) holds if $\A$ has a intersection stable, countable generating set.
	If $\T$ is (at most) countable, then (1) and (2) hold. In this case every probability measure $P$ on $\X \times \T$ gives rise to a drift process. 
	
	\subsection{Drift as change of distribution}
	
	In this subsection, we discuss that Definition \ref{def2} can be simplified 
	to the fact that  the probability 
	distribution is constant, i.e.\   $p_t$ does not depend on time (up to a null set). This is the standard setting of classical (drift free) machine learning.
	
	\begin{definition}
		Let $(p_t,P_T)$ be a drift process. We say that $p_t$ is \textdef{constant} if there exists a probability measure $P_X \in \Prob(\X)$ such that $p_t = P_X$ for $P_T$-a.s. all $t \in \T$.
		We say that $p_t$ has a \textdef{change of distribution} or is changing iff it is not the case that $p_t$ is constant.
	\end{definition}
	
	It is clear that $P_X$ is uniquely determined by this property. 
	Furthermore we can show that $P_X$ given $p_t$ is characterized by the expectation with respect to $P_T$ (Corollary~\ref{ConstantDriftIsExpectation}). Indeed, we can even find a $t_0 \in \T$ such that $P_X = p_{t_0}$, i.e. $P_X$ actually appears at some (actually nearly every) point in time (Lemma~\ref{ConstantDriftIsRealizedAtSomePoint}).
	
	
	This enables us to characterize the relation between drifting processes and change of distribution:\footnote{Notice that we have to take care of the null sets here.}
	
	\begin{theorem}
		\label{ProbabilisticDriftAndConstantDriftAreEquivalent}
		Let $(p_t,P_T)$ be a drift process. Then $p_t$ is constant if and only if $p_t$ has no drift.
		%
	\end{theorem}

	Though it does not hold that drift processes with no proper drift are constant, it can be proven, that a drift process has no proper drift if an only if ${p_t \otimes P_T = P_X \times P_T}$\footnote{See Definition~\ref{def:ProductMeasure} for details}, for some probability measure $P_X$ (Lemma~\ref{KeyCharacterizationOfProperDrift}). 
	
	\subsection{Drift as change of model}
	\label{SecModelDrift}
	
	We will now consider drift in the context of machine learning models; machine learning models in the context of drift often learn a constant model over a time window. It is common practice to detect drift by a change of such model,  e.g.\ a changed error. Here, we are not interested in specific models, rather we consider  
	$\T$-invariant models, which we will use as prototypical optimum machine learning model: 
	\begin{definition}
		Let $(p_t,P_T)$ be a drift process. For a $P_T$-non-null set $A \in \B$ we define the \textdef{$\T$-invariant model} of $p_t$ over $A$ as the marginalization of ${(p_t \otimes P_T)( \cdot \mid \X \times A )}$ onto $\X$ or equivalent \[p_A := \frac{1}{P_T(A)} \int_A p_t \d P_T 
		.\] 
	\end{definition}
	$p_A$ is the optimal, time invariant model in the sense that every static probabilistic  model  that is capable of universal approximations, trained with data observed during $A$ only, converges to $p_A$. 
	
	Furthermore notice that those models have some Bayesian-like properties:  for disjoint non-null sets $A,B \in \B$ it holds \[p_{A \cup B} = \frac{1}{P_T(A) + P_T(B)}(P_T(A)p_A + P_T(B)p_B).\] 
	
	Now we can characterize  drift in terms of models derived from the drift process not being constant:
	\begin{definition}
		We say that a pair of $P_T$-non-null sets $A,B \in \B$ are \textdef{alternating sets} iff $p_A \neq p_B$. 
		If alternating sets exist, then we say that $p_t$ has \textdef{model drift}.
	\end{definition}
	
	Model drift characterizes the fact that an optimal model for observed streaming data, changes over time.
	Having in mind that a practical model approximates  the 
	behavior of an optimal $\T$-invariant model, we see that model drift captures common practice: e.g.\ many drift detection methods refer to a change in model accuracy  \cite{adwin,ddm,eddm}.
	
	We will now investigate the relation of model drift and proper drift:
	
	\begin{theorem}
		\label{ModelDriftAndProperDriftAreEquivalent}
		Let $(p_t,P_T)$ be a drift process and let $\B_0 \subset \B$ a generating set (i.e.\ $\sigma(\B_0) = \B$\;\footnote{See Definition~\ref{def:GeneratorOfSigmaAlgebra} for details}), which is stable under finite intersections. Then the following properties are equivalent:
		\\\noindent\hspace*{4mm} 1. $p_t$ has proper drift,
		\\\noindent\hspace*{4mm} 2. $p_t$ has model drift,
		\\\noindent\hspace*{4mm} 3. there exist alternating sets $A,A^C$, with $A \in \B_0$.
	\end{theorem}
	
	Besides the observation that model drift is equivalent to proper drift, Theorem~\ref{ModelDriftAndProperDriftAreEquivalent} has an interesting consequence regarding the structure of alternating sets: alternating sets  take  the form of complementary subsets of $\T$. 
	Provided the index set $\T$ represents time, i.e.\ is contained in the real numbers, this  implies that 
	model drift is the same as the   existence of a change-point:
	
	\begin{corollary}
		\label{Changepoint}
		Let $(p_t,P_T)$ be a drift process and suppose that $\T \in \{ [a,b],\R_{\geq 0},\R\}, \B = \Borel(\T)$. If $p_t$ has proper drift, then there exists a change-point, i.e. it exists a $t_0 \in \R$ such that $p_{\{t < t_0\}} \neq p_{\{t \geq t_0\}}$.
	\end{corollary}
	
	This result can be seen as a justification of change-point detection methods in the field of drift detection.
	
	\subsection{Drift as non-stationarity of a stochastic process}
	
	In the context of time-series, the notion of  stationary processes constitutes a prominent concept \cite{Fundamentalsofprobabilityandstochasticprocesseswithapplicationstocommunications}. We discuss its relation to drift. In this section let $\T \in \{\N,\Q,\R\}$, so that we have a natural shift operation $\cdot + \tau$ with $\tau \in \T$, representing shift in time.
	
	\begin{definition}
		Let $(\Omega,\mathcal{F},\P)$ be a probability space. A stochastic process $X_t : \Omega \times \T \to \X$ is \textdef{stationary} if \[\P \circ (X_{t_1},...,X_{t_n})^{-1} = \P \circ (X_{t_1+\tau},...,X_{t_n+\tau})^{-1}\] for all $t_1,...,t_n \in \T$, $\tau \in \T$ and $n \in \N$. 
	\end{definition}
	
	Notice that stationary implies having no drift.
	
	\begin{theorem}
		\label{Stationary}
		Let $X_t$ be a stochastic process. For every sequence $t_1,...,t_n \in \T$ we obtain a non-probabilistic drift process by setting ${p^{(t_1,...,t_n)}_\tau = \P \circ (X_{t_1+\tau},...,X_{t_n+\tau})^{-1}}$. 
		If $X_t$ is stationary then $(p^{(t_1,...,t_n)}_\tau,P_T)$ has no drift for all probability measures $P_T$ on $\T$, $t_1,...,t_n \in \T$ and $n \in \N$.\\
		Furthermore, if $P_T$ has no null sets, then the notion of stationarity of $X_t$ and $(p^{(t_1,...,t_n)}_\tau,P_T)$ having no drift for all $t_1,...,t_n \in \T$ and $n \in \N$, are equivalent. 
	\end{theorem}
	
	
	The reason why having no drift does not  imply stationarity comes from the fact that the latter is defined point-wise for all $\tau \in \T$. It is therefore  equivalent to the non-probabilistic definition of no drift \cite{asurveyonconceptdriftadaption}, using the same transformation as used in Theorem~\ref{Stationary}. However additional assumptions could be added to induce such implication, e.g.\ by assuming that $X_t$ has $\P$-a.s. (Corollary~\ref{thm:ContinuousPathes})\ continuous paths\footnote{See Definition~\ref{def:StochProcess}}. 
	
	\subsection{Drift as dependency between data and time}
	\label{SecDriftAsDependency}
	
	In addition to these notions of drift from the literature,
	we will now discuss drift under a novel aspect, which will be particularly suited to derive efficient algorithms, namely in the context of independence of random variables.
	
	In the classical machine learning setup one considers samples as realizations of (independent) identically distributed $\X$-valued random variables. In the context of drift,  this distribution changes, as discussed above.
	To put this into the context of  dependence of variables,
	we can equip each sample with a timestamp of its occurrence: instead of  $\X$-valued random variables $X$, we  consider $\X \times \T$-valued random variables $(X,T)$. If there is no drift then the distribution of $X$ should not depend on $T$, i.e.\ $X$ and $T$ should be independent: 
	
	\begin{definition}
		Let $(p_t,P_T)$ be a drift process and let $(X,T) \sim p_t \otimes P_T$ a pair of random variables. We say that $p_t$ has \textdef{dependency drift} if $X$ and $T$ are not independent. 
	\end{definition}
	
	It turns out that this is an alternative characterization of proper drift:
	
	\begin{theorem}
		\label{DriftAsDependency}
		Let $(p_t, P_T)$ be a drift process. Then $p_t$ has proper drift if and only if it has dependency drift.
	\end{theorem}
	
	This result allows us to reduce the problem of drift detection to the problem to test independence of variables. The latter problem is well investigated and highly efficient algorithms exist for independence tests.
	
	\section{APPLICATIONS}
	
	In Theorem~\ref{DriftAsDependency} we showed that drift can be described as the dependency between data and time. We will now make use of this by construction two methods: A fast, ADWIN~\cite{adwin} based drift detector (SWIDD) and a drift explanation method (DriFDA).
	
	\subsection{Single Window Independence Drift Detection (SWIDD)}
	
	Drift detection refers  to the task to determine whether there is a change in an observed data stream.
	Most drift detection methods \cite{adwin,ddm,eddm,pagehinkley,entropybaseddriftdetection} detect drift using a two window approach; samples are hold in two windows that are assumed to be sampled from the same base distribution, so that drift may be detected using a two-sample test. This may be done directly as in \cite{pagehinkley} or after a transformation, i.e.\ use the prediction error of a  model  \cite{adwin,ddm,eddm}. 
	
	We will rely on the pipeline as proposed in ADWIN as  particularly popular method.
	ADWIN~\cite{adwin} stores the incoming prediction errors  in a sliding, size-adaptive window that is successively split into two windows. Those two windows are then tested against each other by checking whether the absolute difference of the mean prediction error exceeds a predefined threshold. If so, a drift is indicated and all samples before that time are discarded.

	We use Theorem~\ref{DriftAsDependency} to extend this idea to detect general drift using  a single window only: Instead of splitting our window we assign every sample with a time-stamp and apply a statistical test to determine whether time and data are dependent or not. This leads to Algorithm~\ref{SWIDDAlg}. Notice that this differs from the usual ADWIN only in line~\ref{line:difftoadwintime} where we add the timestamp to $x_i$ of the moment of its arrival, rather than the prediction error and in line~\ref{line:difftoadwintest} where we use an independence test to check for drift 
	(Theorem~\ref{DriftAsDependency}), rather than window splitting. 
	We implement
	\footnote{The implementations of our proposed methods are available on GitHub - \url{https://github.com/FabianHinder/drifting-data-in-continuous-time}}%
	SWIDD (Algorithm~\ref{SWIDDAlg}) based on the Hilbert-Schmidt Independence Criterion (HSIC)~\cite{hsic}.
	
	\begin{algorithm}
		\caption{SWIDD}
		\label{SWIDDAlg}
		\begin{algorithmic}[1]
			\Procedure{SWIDD: Single Window Independence Drift Detector}{$(x_i)$ data stream, $p$ $p$-value for statistical test, $n_{\min}$ minimal number of samples in window} \;
			\State Initialize Window $W \gets []$\;
			\While{Not at end of stream $(x_i)$} \;
			\State $W \gets W \cup \{ (x_i,t_i) \}$ \; \label{line:difftoadwintime}
			\Comment{i.e. add new sample $x_i$ received at time $t_i$}
			\Repeat{ Drop element from the tail of $W$}
			\Until{$|W| < n_{\min}$ or $\Call{Test}{W,p}$ accepts $H_0$} \label{line:difftoadwintest}
			\EndWhile
			\EndProcedure
		\end{algorithmic}
	\end{algorithm}
	
	A beneficial property of SWIDD is that we may use a test for general or conditional independence; 
	which allows us to apply it to virtual and real drift alike. Furthermore, since we do not depend on a model, we are not subject to its specific deficiencies (see Figure~\ref{fig:data0drifted}). 
	
	SWIDD is superior to a fixed two-window approach when it comes to continuous, in particular fast and periodic drift. This is caused by the fact that two window approaches assume an identical distribution at least for a single window; a counter example would be $p_t = \mathcal{N}( \sin(t) , \sigma)$ and a window size of $2n\pi$, where $\mathcal{N}(\mu,\sigma)$ denotes the normal distribution. Though this problem can be solved by dynamic window selection as used by ADWIN (see Corollary~\ref{Changepoint}), it causes a considerable amount of computation time.

	
	\subsubsection{Experiments}
	To demonstrate the generality of our approach, we created two artificial data sets which highlight
	typical challenges of existing approaches for drift detection. In addition, we compare SWIDD to the  drift detection methods ADWIN~\cite{adwin} and DDM~\cite{ddm}, which depend on the classification accuracy of a supervised model, as well as against the (unsupervised) statistical methods HDDDM~\cite{hdddm} and HDDDM where we replaced the Hellinger-distance and the t-test by the kernel two sample test~\cite{kernel2sampletest} (referred to as K2ST), on the SEA data set~\cite{seadata} and the rotating hyperplane data set~\cite{rotatinghyperplanedata} (referred to as RPLANE).
	
	\begin{figure}[b]
		\centering
		\begin{minipage}[b]{0.22\textwidth}
			\includegraphics[width=\textwidth]{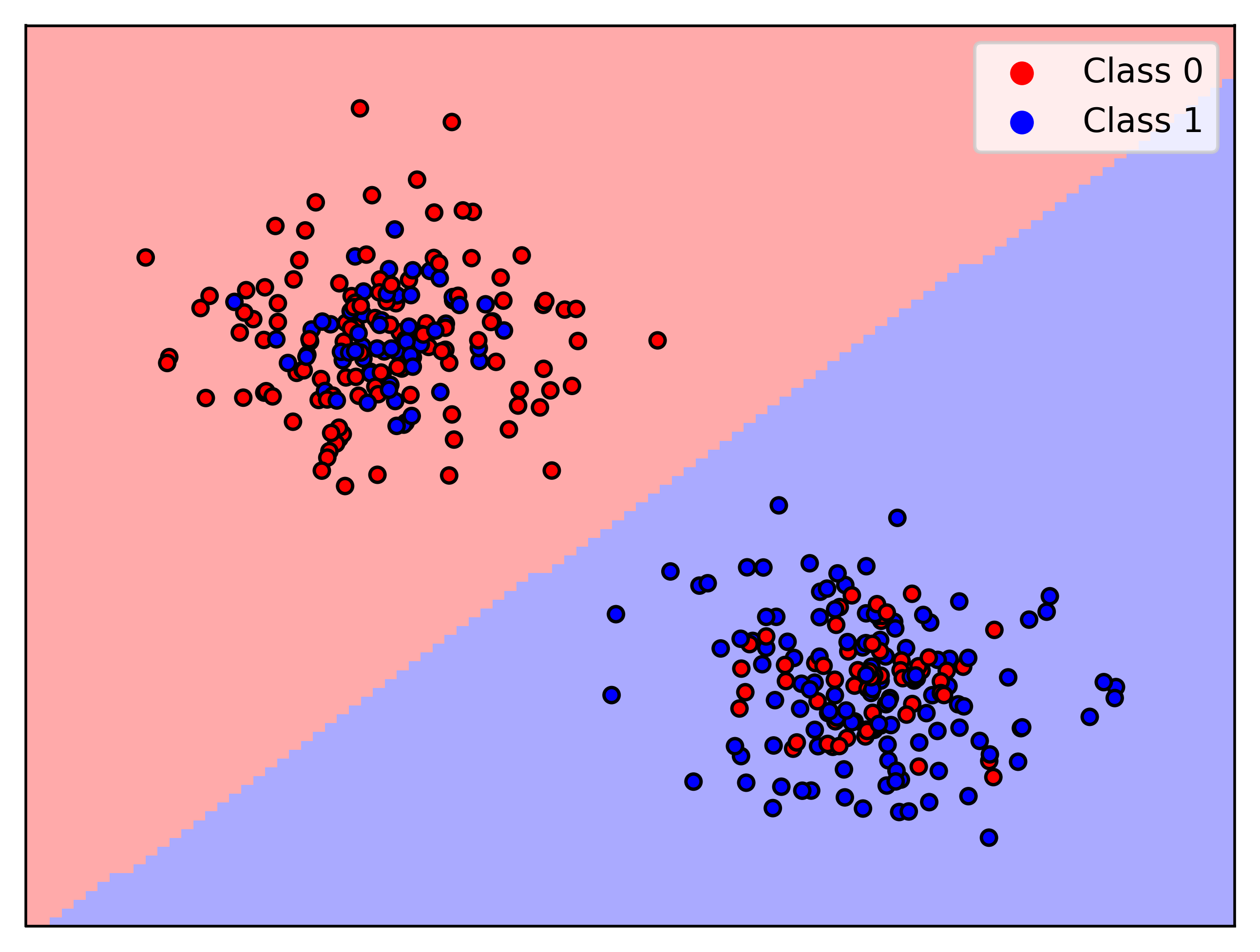}
			\subcaption{{A linear classifier fitted to the original data set.\\}\label{fig:data0}}
		\end{minipage}
		\hfill
		\begin{minipage}[b]{0.22\textwidth}
			\includegraphics[width=\textwidth]{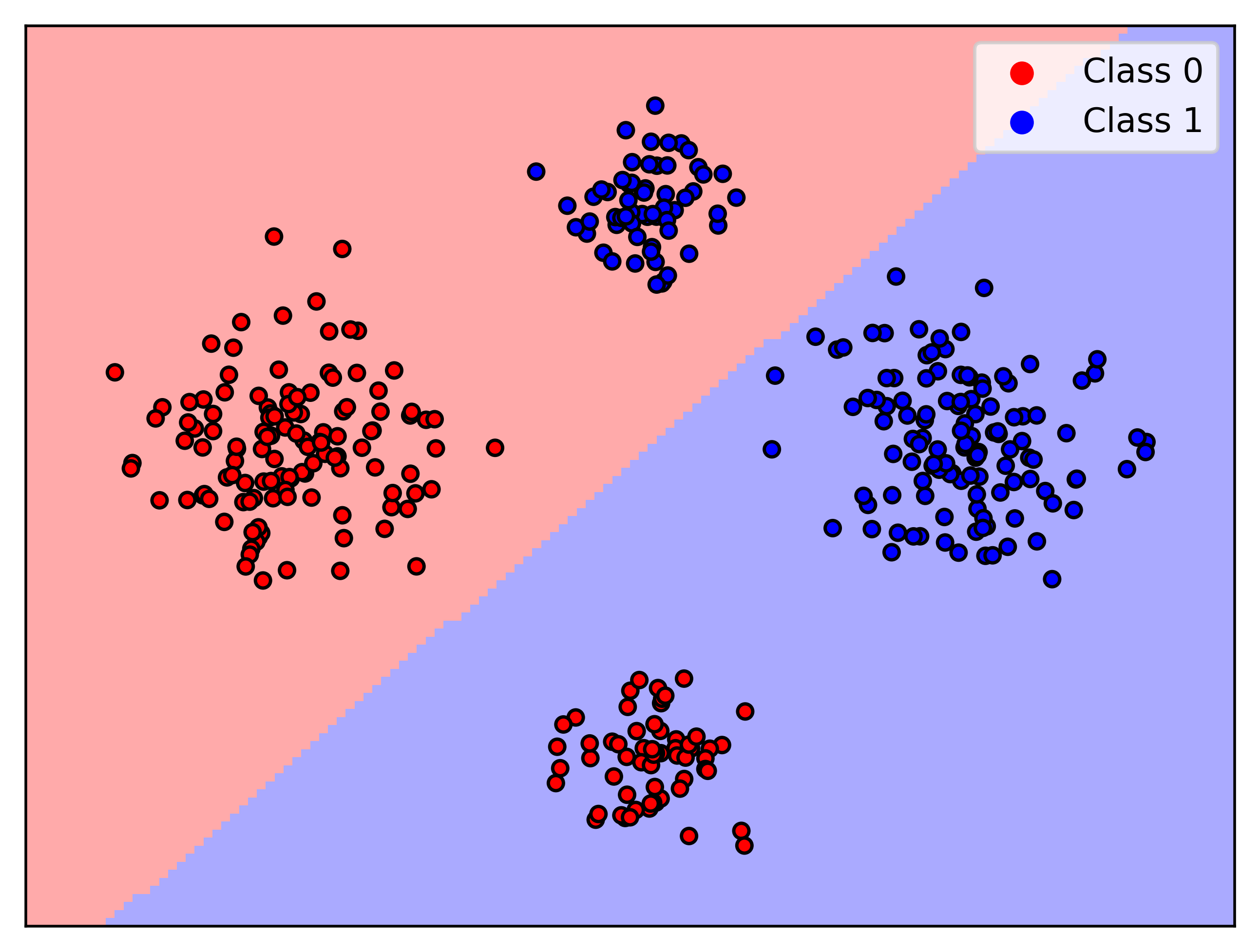}
			\subcaption{{Drifted data set. Note that the accuracy does not change.}\label{fig:data0drifted}}
		\end{minipage}
		\caption{Fooling error based drift detection methods}
	\end{figure}
	
	\paragraph*{Comparison to supervised drift detection}
	Methods such as 
	ADWIN~\cite{adwin}, DDM~\cite{ddm} and EDDM~\cite{eddm}
	use the classification error as an indicator of drift. They assume that drift leads to a change (e.g. decrease) in accuracy. We construct a scenario in which this assumption does not hold: We create a binary classification data set with two clusters.
	Both clusters are mixtures of samples from both classes, but with different dominance.
	A linear classifier yields a decision boundary as shown in Figure~\ref{fig:data0}. Drift is constructed by moving all samples from one class along the decision boundary in the direction of the upper right corner, whereby we do not cross the decision boundary. The final  scenario is shown in Figure~\ref{fig:data0drifted}.  Error-based drift detectors do not detect this drift because the classification error does not change when moving the data points this way, unless the classifier is retrained.
	It would be possible to obtain a better (in the limit perfect) accuracy; yet, active drift learners would require a drift detection to do so. SWIDD detects this drift since it does not rely on the classification error.
	
	\begin{figure}[t]
		\centering
		\begin{minipage}[b]{0.22\textwidth}
			\includegraphics[width=\textwidth]{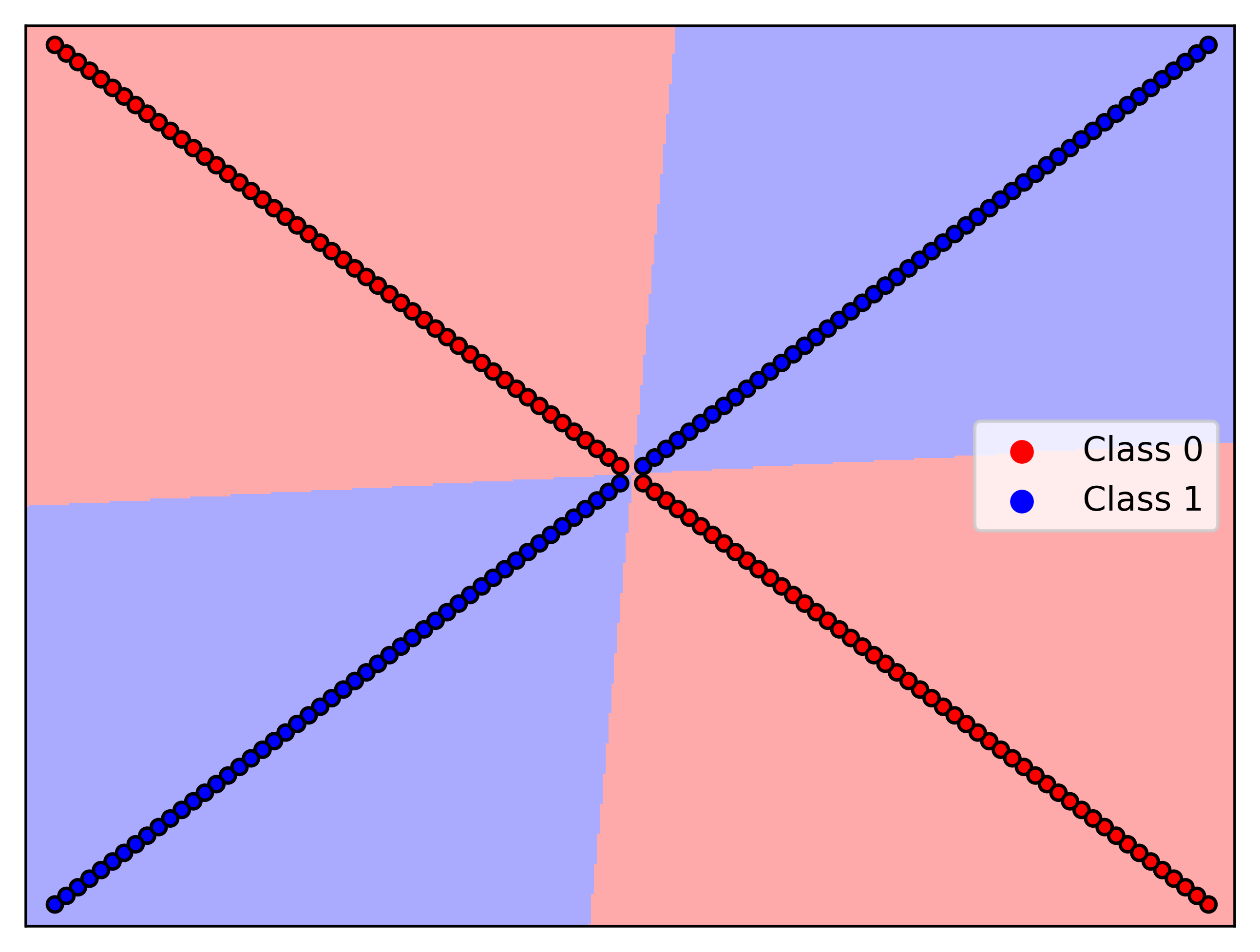}
			\subcaption{{A quadratic discriminant model fitted to the original data set.\\}\label{fig:data1}}
		\end{minipage}
		\hfill
		\begin{minipage}[b]{0.22\textwidth}
			\includegraphics[width=\textwidth]{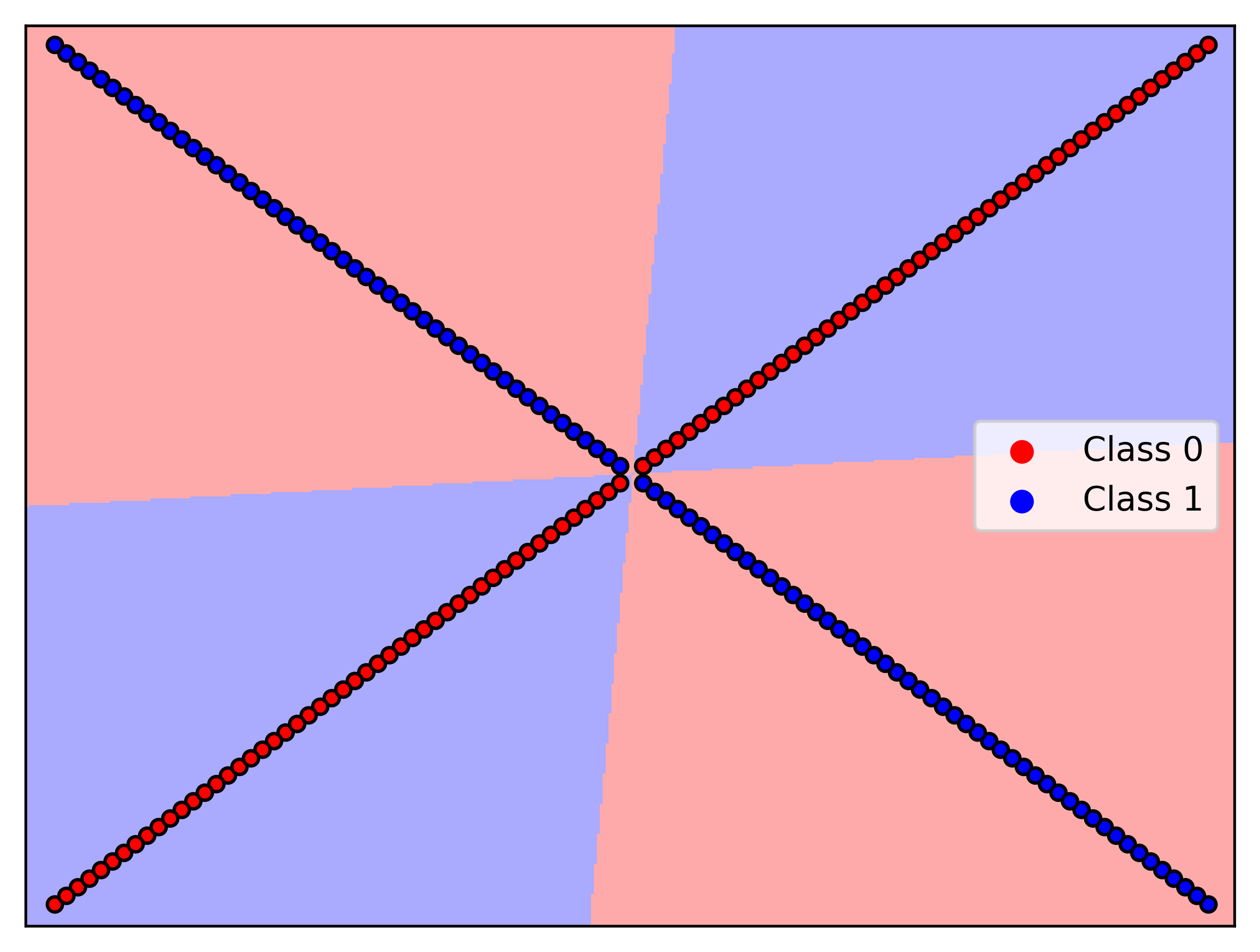}
			\subcaption{{Drifted data set. Mean, variance and feature-wise marginals did not change.}    \label{fig:data1drifted}}
		\end{minipage}
		\caption{Fooling simple distributional drift detectors}
	\end{figure}
	
	\paragraph*{Comparison to unsupervised drift detection}
	Another class of methods for drift detection is based on distributional changes \cite{detectingchangesindatastreams, nonparametricchangepointdetection, detectingchangesintimeseries, entropybaseddriftdetection, hdddm, informationtheoreticchangedetection, statchangedetectionmultidimdata, kernel2sampletest}. These methods try to detect drift by detecting changes in the sampling distribution of the data stream. Many of these methods~\cite{detectingchangesindatastreams, nonparametricchangepointdetection, detectingchangesintimeseries, entropybaseddriftdetection, hdddm} use some kind of windowing - split the data stream (or parts of it) into two windows and compute statistics on these windows. However, relying on two windows can be problematic because we have to select the right length of the window so that quickly occurring abrupt drifts are recognized -- usually, it is assumed that the distribution of the samples in a window is fixed. Another problem of some of these methods is that they try to reduce computational complexity by assuming that the drift will show up in the mean, variance or feature-wise marginals~\cite{hdddm,entropybaseddriftdetection}. This is problematic because one can construct drifting data sets where the mean, variance and the feature-wise marginal distribution do not change - such drifts can not be perceived by methods that make these simplifying assumptions. For instance we can construct a data set where the points are arranged like a cross so that each class has its own diagonal - see Figure~\ref{fig:data1}. If the cross is symmetric and if the samples are placed symmetrically around the center, then we can swap the labels of the two diagonals - see Figure~\ref{fig:data1drifted} - but the mean, variance and the feature-wise marginal distributions do not change. Therefore, these methods do not recognize the drift. However, our method is able to detect this drift since it does not make any simplifying assumptions about the distributional changes.
	
	\paragraph*{Benchmarks}
	We compare SWIDD on common benchmark data sets:  the SEA data set~\cite{seadata} and the rotating hyperplane data set~\cite{rotatinghyperplanedata}. We recorded the mean F1-score and mean computation time - over three runs with different random seeds - and compared SWIDD to HDDDM, K2ST, ADWIN and DDM. 
	The results are shown in Table~\ref{tab:comparedriftdetectors}.  SWIDD is best for SEA and second best for RPLANE, being reasonably fast in both cases.
	\begin{table}[b]
		\centering
		\caption{Mean F1-score and mean computation time of drift detectors on two common benchmark data sets.}
		\begin{tabular}{|c||c|c||c|c|}
			\hline
			& \multicolumn{2}{|c||}{Data set} & \multicolumn{2}{|c|}{Computation time} \\
			\hline
			Method & RPLANE & SEA & RPLANE & SEA \\
			\hline\hline
			SWIDD & $0.86$ & $0.63$ & $2.59$s & $3.61$s \\ \hline
			HDDDM & $0.44$ & $0.44$ & $0.02$s & $0.03$s \\ \hline
			K2ST  & $1.00$ & $0.22$ & $5.74$s & $19.87$s \\ \hline
			ADWIN & $0.60$ & $0.33$ & $0.12$s & $0.14$s \\ \hline
			DDM   & $0.70$ & $0.13$ & $0.06$s & $0.06$s \\ \hline
		\end{tabular}
		\label{tab:comparedriftdetectors}
	\end{table}
	
	\subsection{Drifting Feature Decomposition Analysis}
	
	\newcommand{\DFDA}{DriFDA}
	\renewcommand{\S}{\textbf{S}}
	Now we aim for a drift explanation method, i.e\ a technology which has the potential to uncover potentially semantically meaningful components from given data. Drifting Feature Decomposition Analysis (\DFDA) aims for a decomposition of the observed data $X$ into a drifting part $X_D$ and a non-drifting part $X_I$.

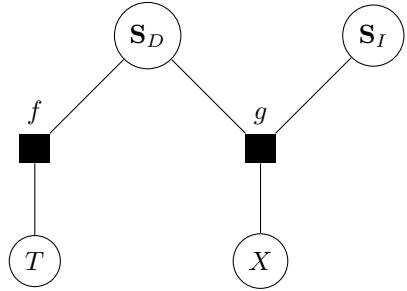
\begin{figure}[!tb]
		\centering
		\begin{tikzpicture}
		[scale=1.5,auto=left,every node]
		\node[circle,draw] (n1) at (1,2) {$\S_D$};
		\node[circle,draw] (n2) at (3,2) {$\S_I$};
		\node[circle,draw] (n5) at (0,0) {$T$};
		\node[circle,draw] (n6) at (2,0) {$X$};
		
		\node[rectangle,fill=black,label={$f$}] (n3) at (0,1) {o};
		\node[rectangle,fill=black,label={$g$}] (n4) at (2,1) {o};
		
		\foreach \from/\to in {n1/n3,n1/n4,n2/n4,n3/n5,n4/n6}
		\draw (\from) -- (\to);
		\end{tikzpicture}
		\caption{DriFDA factor graph}
		\label{fig:DriFDAFactorGraph}
	\end{figure}

	Let $(X,T) \sim p_t \otimes P_T$ with $\X = \R^d$ and $\T = \R_{\geq 0}$. By Theorem~\ref{DriftAsDependency} drift is the same as dependency between $X$ and $T$. If we model our data using independent, hidden source variables $\S_D$ and $\S_I$ that determine $X$ and $T$, i.e.\ $f(\S_D) = T$ and $g(\S_D,\S_I) = X$, we arrive at the factor graph presented in Figure~\ref{fig:DriFDAFactorGraph}.

	Therefore it is reasonable to define $X_D$ resp.\ $X_I$ as the best possible approximation of $X$ using the information encoded in $\S_D$ resp.\ $\S_I$ only. In mathematical terms, we may express this idea using the notion of conditional expectation, i.e.\ we define
	\begin{align*}
	X_D &:= \E[g(\S_D,\S_I) \mid \S_D ] \\
	X_I &:= \E[g(\S_D,\S_I) \mid \S_I ].
	\end{align*}
	Since $\S_D$ and $\S_I$ are assumed to be independent and $\S_D$ determines $T$ it follows that $X_I$ must be independent of $T$ and therefore it can not have drift. This on the other hand implies that $X_D$ has to contain the entire drift information of $X$. Now by minimizing the information of $\S_D$ or maximizing the information of $\S_I$ (this depends on the chosen model), we force $\S_I$, and therefore $X_I$, to contain the entire non-drifting information of $X$.
	Concrete methods depend on the choice of the functional form of $f$ and $g$.
	
	
	\begin{figure}[!tb]
		\centering
		\includegraphics[width=0.5\textwidth]{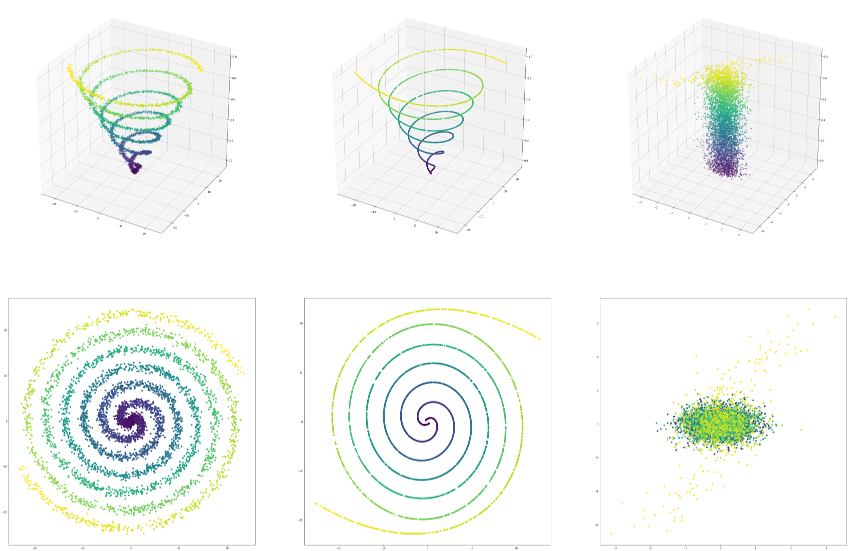}
		\caption{$k$-curve-DriFDA applied to twister data set. Original data ($X$), drift component ($X_D$), decomposition error ($X-X_D$) (Time is displayed as color and on $Z$-axis in upper row). }
	\end{figure}
	
	\subsubsection{Linear-DriFDA}
	
	A first approach to implement this method is by assuming that $f$ and $g$ are linear. Instead of estimating $f$, $g$, $\S_D$ and $\S_I$ all separately we may combine $\S_D$ and $\S_I$ resp.\ $f$ and $g$ into a single vector $\S$ resp.\ a single map represented by a matrix $A$. 
	Under those assumptions we can compute $X_D$ and $X_I$:
	\begin{lemma}
		\label{LinearDriFDAComputation}
		In the situation described above it holds \begin{align*}X_D &= A \S_D + A\E[\S_I],\\X_I &= A \S_I + A\E[\S_D],\\X_D + X_I &= X + \E[X].\end{align*} 
	\end{lemma}
	
	Instead of forcing $A$ to have a specific shape, we may simply train our model for a general linear form and apply feature selection, with respect to $T$, to determine whether a specific component of $\S$ belongs to $\S_D$ or $\S_I$. To assure that we can do this component-wise we need to assume the components of $\S$ to be independent. Note that this renders mutual information a particularly good choice as feature selection strategy, mirroring the assumed independence, i.e.\ non-redundancy of features  \cite{Featureselectionbasedonmutualinformationcriteriaofmaxdependencymaxrelevanceandminredundance}. 
	
	To determine $A$ and $\S$ we can use an independent component analysis (ICA) \cite{ICA} to $(X,T)$ - this leads to Algorithm~\ref{alg:LinearDriFDAAlg}.
	
	\begin{algorithm}
		\begin{algorithmic}[1]
			\Procedure{Linear-DriFDA: Drifting Feature Decomposition Analysis under linearity assumption}{$(X,T) = (x(j),t(j))_{j=1...N}$ data steam, $n$ number of independent blind-sources, $I_{\min}$ minimal dependency} \;
			\State $(\S,A) \gets \Call{ICA}{(X,T),n}$\;
			\For{$i \in \{1,...,n\}$}
			\State $I_i \gets I(S_i,T)$ \;
			\Comment{Compute mutual information}
			\If{$I_i \geq I_{\min}$}
			\State $(S_D)_i \gets S_i$ 
			\Else
			\State $(S_D)_i \gets \frac{1}{N}\sum_{j = 1}^N S_i(j)$ \; 
			\Comment{Mean value}
			\EndIf
			\EndFor
			\State $\textbf{return } A \S_D$ \;
			\Comment{Invert decomposition with drift relevant sources only}
			\EndProcedure
		\end{algorithmic}
		\caption{Linear-DriFDA}
		\label{alg:LinearDriFDAAlg}
	\end{algorithm}
	
	
	\subsubsection{{\it k} -curve-DriFDA}
	
	We model $p_t$ as a mixture of drifting Gaussians, i.e.\ \[p_t = \sum_{i = 1}^n \lambda_i(t) \mathcal{N}(\mu_i(t), \sigma_i(t))\] with $0 \leq \lambda_i(t)$, $\sum_i \lambda_i(t) = 1$ for all $t$. Then we can implement DriFDA as a generalized Gaussian-mixture clustering, where we estimate $k$ curves that correspond to the means and variances. 
	Under this assumption we may construct our model as follows: We choose
	\begin{align*}
	\S_D &= (t,i)
	\end{align*}
	where $t$ is the time and $i$ corresponds to the Gaussian generating the sample. It is natural to define 
	\begin{align*}
	g( (t,i) , \S_I  ) := \sigma_i(t) \S_I + \mu_i(t).
	\end{align*}
	So $\S_I$ generates row samples, which  are then shaped using $\S_D$.
	Then it holds 
	\begin{align*}
	X_D(t,i) &= \mu_i(t) + \sigma_i(t)\E[\S_I], \\ \S_I \mid (t,i),X &= \frac{1}{\sigma_i(t)}(X-\mu_i(t)).
	\end{align*}
	This implies that adapting $\sigma_i$ and $\mu_i$ corresponds to maximizing the Gaussianity, and therefore information, of $\S_I$ (cf. \citet{ICA}). Indeed, if $\sigma_i$ and $\mu_i$ are adapted perfectly then $\S_I \sim \mathcal{N}(0,1)$, which then implies that $X_D(t,i) = \mu_i(t)$.
	
	
	We may approximate the Gaussian clustering by a $k$-means algorithm for simplicity (cf. \citet{Bishop}), i.e.\ we set $i = \argmin_i \Vert \mu_i(t) - x \Vert$. The obtained algorithm can be found in the supplemental material (Algorithm~\ref{kCDriFDAAlg}). Notice that an incremental insertion of the data points may be used, rather than inserting them all at once, to reduce unwanted jumping behaviors of the mean-value-curves.
	
	
	\subsubsection{Experiments}
	
	We applied our two DriFDA variants to various artificial and real-world data sets. Since we want to decompose our data ($X$) into a drifting ($X_D$) and a non-drifting part we may quantify the reliability of our methods by measuring the dependency between the decomposition error $X-X_D$ and time $T$. To do so we use the prediction error with time as objective value and a $k$-nearest-neighbors model; the error on $X$ serves as a baseline.  For $k$-curves-DriFDA we used $k=20$ RBF-networks with $10$ prototypes, data was presented in $20$ chunks. For Linear-DriFDA we used the mean overall mutual information as threshold. We use the Airlines, Electricity and Poker-Hand data sets from the MOA data set repository \cite{MOA}. The artificial data sets \emph{twister}: $p_t = \mathcal{N}( \alpha t (\sin(\beta t),\cos(\beta t)) , \sigma )$, \emph{spiral}: $p_t = \mathcal{N}( \alpha (\sin(\beta t),\cos(\beta t)) , \sigma )$, \emph{Y}: $p_t = \frac{1}{2}(\delta_{\max(0,t-\alpha)}+\delta_{-\max(0,t-\alpha)}) \times \mathcal{U}([0,1])$ and \emph{square}: $p_t = \delta_{(\alpha t, \beta t)}*\mathcal{U}^2([0,1])$ were designed to provide ground truth, here $\delta$ denotes the Dirac measure and $\mathcal{U}$  the uniform measure.

	\begin{table}[!tb]
		\centering
		\caption{Mean and variance (if $\geq 0.01$) over 8 runs. $*$ no chunk wise adaption was used; $\dagger$ 40 chunks and $k=4$ curves were used. }
		\begin{tabular}{|c||c|c|}
			\hline
			Data set & $k$-curves-DriFDA & Lin.-DriFDA \\
			\hline \hline
			Airlines&$0.67^*$&$0.92 (\pm0.01)$ \\ \hline
			Electricity&$0.54$&$0.70 (\pm0.01)$ \\ \hline
			Poker-Hand&$0.21$&$0.22 (\pm0.01)$ \\ \hline \hline
			twister&$0.13^\dagger$&$0.92$ \\ \hline
			spiral&$0.02^*$&$0.13$ \\ \hline
			Y&$0.06^*$&$0.02$ \\ \hline
			square&$0.04^*$&$0.02$ \\ \hline
		\end{tabular}
		\label{tab:resultsdrifda}
	\end{table}
	
	Results are displayed in Table~\ref{tab:resultsdrifda}.
	Though Linear-DriFDA is only capable of finding linear relationships it works surprisingly well on a large fraction of the data sets. 
	For $k$-curves-DriFDA, some results are excellent. The number of chunks used to present the data  seems to be a very relevant hyper-parameter, hence an automatic optimization scheme  or a robust selection technology would be helpful.
	
	
	\section{DISCUSSION}
	We have presented formal definitions of drift in continuous time, this way substantiating common practice such as learning on time windows, drift detection by referring to model errors, or change point detection by 
	a mathematical justification. In addition, we derived a particularly elegant novel characterization in terms of independence of observations and time, which opens the way towards efficient and flexible algorithms which are based on classical independence tests. We have demonstrated this potential by a novel drift detection method, and a novel decomposition method which can disentangle drifting and non-drifting part of observed signals. The latter has so far been tested in first benchmarks only, displaying a robust and surprisingly efficient behavior. The  suitability to uncover semantically meaningful signals in the context of larger applications and specific domain expertise is subject of ongoing work. 
	
	\subsection*{Acknowledgement}
	Funding by the VW Foundation in the frame of the project IMPACT, and by the BMBF for the project ITS\_ML, grant number 01IS18041, is gratefully acknowledged.  
	
	\bibliography{bib}
	
	\newpage
	
\renewcommand\thesection{\Alph{section}}
\setcounter{section}{0}
\section{SUPPLEMENTAL MATERIAL}

\theoremstyle{definition}
\newtheorem{definition_}{Definition}
\theoremstyle{plain}
\newtheorem{theorem_}{Theorem}
\newtheorem{lemma_}{Lemma}
\newtheorem{corollary_}{Corollary}

\theoremstyle{definition}
\newtheorem{definition__}{Definition}[section]

\theoremstyle{plain}
\newtheorem{lemma__}{Lemma}[section]
\newtheorem{theorem__}{Theorem}[section]
\newtheorem{corollary__}{Corollary}[section]

\theoremstyle{remark}
\newtheorem{remark__}{Remark}[section]

\subsection{Theorems and proofs}
We will now give additional definition, remarks, theorems, lemmas and corollaries. In particular we will provide proofs for the theorems given in the paper. Note that we will include the definitions and theorems given in the paper using the same numeration as before. 
\subsubsection{Definition of a drift process}
\begin{definition__}
	\label{def:MarkovKernel}
	Let $(\T,\B),(\X,\A)$ be two measurable spaces. A \textdef{Markov kernel} is a map $\kappa : \T \times \A \to \R$ such that:
	\begin{enumerate}
		\item $t \mapsto \kappa(t,A)$ is measurable for all $A \in \A$,
		\item $\kappa(t, \cdot )$ is a probability measure for all $t \in \T$.
	\end{enumerate}
\end{definition__}

\begin{definition_}
	Let $(\T,\B)$ and $(\X,\A)$ be two measurable spaces.
	A \textdef{drift process} $(p_t,P_T)$ is a Markov kernel $p_t$ from $\T$ to $\X$  and a probability measure $P_T$ on $\T$.  
\end{definition_}

\begin{remark__}
	\label{thm:MarkovIsNatural}
	Notice that
	\begin{enumerate}
		\item Markov kernels are exactly the measurable maps $\kappa : \T \to \Prob(\X)$, where $\Prob(\X)$ is the set of all probability measures on $\X$ equipped with the initial $\sigma$-algebra induced by all evaluation maps ${\Phi_A : P \mapsto P(A)}$ for $A \in \A$.
		\item If we assume that $\T$ is a topological space, then every continuous map $\kappa : \T \to \Prob(\X)$ is a Markov kernel, here we equip $\Prob(\X)$ with the topology induced by the total variation norm. This follows by writing $\Vert P-Q \Vert_\text{TV} = \sup_{A \in \A} |P(A)-Q(A)|$ implying $t \mapsto \kappa(t,A)$ is continuous, and hence measurable, for all $A \in \A$.
	\end{enumerate}
\end{remark__}

\begin{definition__}
	\label{def:GeneratedSigmaAlgebra}
	Let $\X$ be some set and $\A_0 \subset 2^\X$ be a set of subsets of $\X$. Then \textdef{the $\sigma$-algebra generated by $\A_0$}, denoted by $\sigma(\A_0)$, is defined as the (with respect to inclusion) smallest $\sigma$-algebra on $\X$ that contains $\A_0$.
	\begin{remark__}
		It can be shows that
		\begin{align*}
		\sigma(\A_0) &= \bigcap_{\mathcal{S} \in \mathcal{F}(\A_0)} \mathcal{S}, \qquad\qquad \text{where} \\
		\mathcal{F}(\A_0) &= \left\lbrace \mathcal{S} \subset 2^X \left| \substack{\text{$\mathcal{S}$ is $\sigma$-algebra on $\X$} \\ \text{ and } \A_0 \subset \mathcal{S}} \right.\right\rbrace.
		\end{align*}
	\end{remark__}
\end{definition__}

\begin{definition__}
	\label{def:GeneratorOfSigmaAlgebra}
	Let $(\X,\A)$ be a measurable space. We say that $\A_0 \subset \A$ is \textdef{a generator of $\A$} iff $\sigma(\A_0) = \A$. We say that $\A_0$ is \textdef{stable under finite intersections} iff for all $A,B \in \A_0$ it holds $A \cap B \in \A_0$.
\end{definition__}

We will make heavy use of the following, well known theorem:
\begin{theorem__}
	Let $(\X,\A)$ be a measurable space and $P$ and $Q$ be probability measures on $\X$. Let $\A_0 \subset \A$ be a generating set, stable under finite intersections. 
	
	Suppose that $P(A) = Q(A)$ for all $A \in \A_0$, then it holds $P = Q$, i.e. $P(A) = Q(A)$ for all $A \in \A$.
	\begin{proof}
		Well known.
	\end{proof}
\end{theorem__}

\begin{definition__}
	\label{def:ProductMeasure}
	Let $(\T,\B),(\X,\A)$ be two measurable spaces. Let $P_T$ and $P_X$ be probability measures on $\T$ resp. $\X$. We call a probability measure $P$ on $(\T \times \X, \sigma(\B \otimes \A))$, where $\B \otimes \A = \{B \times A \mid A \in \A, B \in \B\}$, such that 
	\begin{align*}
	P_T(B)P_X(A) = P(B \times A)
	\end{align*}
	for all $A \in \A, B \in \B$ the \textdef{product measure} of $P_T$ and $P_X$ and denote it by $P_T \times P_X$.
	\begin{remark__}
		It can be shown that product measures always exist and that they are uniquely determined, justifying the notation above. 
	\end{remark__}
\end{definition__}

\begin{remark__}[Fubini's theorem for Markov kernels]
	\label{thm:FubiniForMarkovKernel}
	Let $(\T,\B),(\X,\A)$ be two measurable spaces. Let $p_t$ be a Markov kernel form $\T$ to $\X$, and $P_T$ a measure on $\T$. There exists a unique probability measure $P$ on $(\T \times \X, \sigma(\B \otimes \A))$, such that
	for all $A \in \A, B \in \B$ it holds
	\begin{align*}
	P(B \times A) &= \int_B p_t(A) P_T(\d t).
	\end{align*}
	We denote this uniquely determined measure by \[p_t \otimes P_T := P.\]
\end{remark__}

\begin{definition_}
	Let $(p_t,P_T)$ be a drift process. We say that $p_t$ has \textdef{no drift} or does not drift if $p_t = p_s$ holds $P_T$-a.s., i.e. $(P_T \times P_T)(\{ (s,t) \in \T\times\T \mid p_t = p_s \}) = 1$. We say that $p_t$ has \textdef{drift} or is drifting if it is not the case that it does not drift.
\end{definition_}

\begin{lemma__}
	\label{ClassicalAndProbabilisticDrift}
	Let $\T$ be a countable index set and $(\X,\A)$ be a measurable space. Then there exists a $\sigma$-algebra $\B$ on $\T$ and a probability measure $P_T$ on $(\T,\B)$ such that for every non-probabilistic drift process $p_t$ it holds: $(p_t,P_T)$ has drift if and only if there exists $t,s \in \T$ such that $p_t \neq p_s$.
	\begin{proof}
		Choose $\B$ as the power set of $\T$ and let ${f : \N \to \T}$ be a counting function. Now define \[P_\N(A) = \frac{6}{\pi^2} \sum_{i \in A} \frac{1}{i^2}\] and $P_T$ as the image measure of $P_\N$ under $f$. Since $P_T$ has no null sets the statement follows.
	\end{proof}
\end{lemma__}

\begin{definition_}
	Let $(p_t,P_T)$ be a drift process. We say that $p_t$ has \textdef{no proper drift} iff there exists a drift process $(p_t',P_T')$, such that $p_t \otimes P_T = p_t' \otimes P_T'$, that does not drift. We say that $p_t$ has \textdef{proper drift} iff it is not the case that it has no proper drift.
\end{definition_}

We make use of the following, well known lemma:
\begin{lemma__}
	\label{thm:UniquenessOfKernel}
	Let $(p_t,P_T), (p_t',P_T')$ be two drift processes. If $\B$ has a countable generating set, stable under finite intersection, then it holds 
	\begin{align*}
	p_t \otimes P_T = p_t' \otimes P_T'
	\end{align*}
	if and only if
	\begin{align*}
	P_T = P_T' \qquad \text{and} \qquad p_t = p_t' \quad P_T-a.s..
	\end{align*}
	\begin{proof} Recall that $P_T,P_T'$ are probability measures and that $p_t,p_t'$ are Markov kernels. Then this is well known (and easy).
	\end{proof}
\end{lemma__}

\subsubsection{Drift as change of distribution}

\begin{definition_}
	Let $(p_t,P_T)$ be a drift process. We say that $p_t$ is \textdef{constant} if there exists a probability measure $P_X \in \Prob(\X)$ such that $p_t = P_X$ for $P_T$-a.s. all $t \in \T$.
	We say that $p_t$ has a \textdef{change of distribution} or is changing iff it is not the case that $p_t$ is constant.
\end{definition_}

\begin{lemma__}
	\label{ConstantDriftIsRealizedAtSomePoint}
	Let $(p_t,P_T)$ be a drift process. Then $p_t$ is constant if and only if there exists a $t_0 \in \T$ such that $p_t = p_{t_0}$ for $P_T$-a.s. all $t \in \T$. In particular we may choose $P_X = p_{t_0}$.
	\begin{proof}
		For "$\Leftarrow$" choose $P_X = p_{t_0}$, for "$\Rightarrow$" note that there exists a $t_0 \in \T$ such that $p_{t_0} = P_X$ and hence $p_{t_0} = p_t$ for $P_T$-a.s. all $t \in \T$.
	\end{proof}
\end{lemma__}

\begin{corollary__}
	\label{ConstantDriftIsExpectation}
	Let $(p_t,P_T)$ be a drift process. Assume that $p_t$ is constant. Then it holds $P_X = \int p_t \d P_T$.
	\begin{proof}
		Let $t_0 \in \T$ as in Lemma~\ref{ConstantDriftIsRealizedAtSomePoint}. Then it holds
		$\int p_t \d P_T = \int p_{t_0} \d P_T = p_{t_0} \int 1 \d P_T = P_X.$
	\end{proof}
\end{corollary__}

We will now give a proof of Theorem~\ref{ProbabilisticDriftAndConstantDriftAreEquivalent_}:
\begin{theorem_}
	\label{ProbabilisticDriftAndConstantDriftAreEquivalent_}
	Let $(p_t,P_T)$ be a drift process. Then $p_t$ is constant if and only if $p_t$ has no drift.
	\begin{proof}
		\textbf{"$\Rightarrow$":} Denote by $C = \{ t | p_t = P_X \}$ and by $D = \{ (t,s) | p_t = p_s \}$. Obviously it holds $C \times C \subset D$. Since $p_t$ is constant we have $P_T(C) = 1$ and hence \[1 = (P_T \times P_T)(C \times C) \leq (P_T \times P_T)(D).\]
		
		\textbf{"$\Leftarrow$":} Since $P_T$ is finite we may write ${(P_T \times P_T) (A) = \int P_T(A^x) P_T(\d x)}$, where $A^x = \{y | (x,y) \in A\}$. This implies that $P_T(\{s \in T | p_s = p_t\}) = 1$ for $P_T$-a.s. all $t \in \T$. Therefore the statement follows by Lemma~\ref{ConstantDriftIsRealizedAtSomePoint}.
	\end{proof}
\end{theorem_}

\begin{lemma__}
	\label{KeyCharacterizationOfProperDrift}
	Let $(p_t, P_T)$ be a drift process. Then $p_t$ has no proper drift if and only if it exists a probability measure $P_X$ such that \[p_t \otimes P_T = P_X \times P_T.\] If $P_X$ exists, then it is unique with this property.
	\begin{proof}
		\textbf{"$\Rightarrow$":} Suppose $p_t$ has no proper drift, let $p_t'$ be the not drifting drift process as in the definition. By Theorem~\ref{ProbabilisticDriftAndConstantDriftAreEquivalent_} there exists a $P_X$ such that \[P_X \times P_T = p_t' \otimes P_T = p_t \otimes P_T.\]
		
		\textbf{"$\Leftarrow$":} We may consider $t \mapsto P_X$ as a constant kernel, i.e. $(P_X,P_T)$ is a drift process. Clearly $P_X$ does not drift and hence $p_t$ has no proper drift.
		
		\textbf{Uniqueness: } for all $A \in \A$ we have \begin{align*}P_X'(A) &= (P_X' \times P_T)(A \times \T) \\&= (P_X \times P_T)(A \times \T) = P_X(A).\end{align*}
	\end{proof}
\end{lemma__}

\subsubsection{Drift as change of model}

\begin{definition_}
	Let $(p_t,P_T)$ be a drift process. For a $P_T$-non-null set $A \in \B$ we define the \textdef{$\T$-invariant model} of $p_t$ over $A$ as the marginalization of $(p_t \otimes P_T)( \cdot | \X \times A )$ onto $\X$ or equivalent \[p_A := \frac{1}{P_T(A)} \int_A p_t \d P_T 
	.\] 
\end{definition_}

\begin{remark__}
	We would like to point out that this notion is by far less theoretical than $p_t$, since single points tend to be null sets, i.e. even with an infinite amount of data, the probability to observe even a single sample at time $t$ is still zero and therefore we cannot estimate $p_t$ directly, even though we have an infinite amount of samples to estimate $p_A$.
	
	In addition notice that (by Corollary~\ref{ConstantDriftIsExpectation}) a drift process is constant if $p_t = p_T$ for $P_T$-a.s. all $t \in \T$, so the notion of model we consider is turned into the classical model, if we assume that no drift takes place.
\end{remark__}

\begin{lemma__}  
	\label{thm:ModelCoveringLemma}
	Let $(p_t,P_T)$ be a drift process and let $A,B,C \in \B$ be pair wise disjoint, non-null sets. 
	Suppose that
	\begin{align*}
	p_A = p_{B \cup C} \qquad \text{and} \qquad p_{A \cup C} = p_B
	\end{align*}
	then it holds $p_A = p_B = p_C$.
	\begin{proof}
		{
			Its a computation: %
			\renewcommand{\A}{P_T(A)}%
			\renewcommand{\B}{P_T(B)}%
			\newcommand{\C}{P_T(C)}%
			\renewcommand{\a}{p_A}%
			\renewcommand{\b}{p_B}%
			\renewcommand{\c}{p_C}%
			Denote by ${\lambda := \frac{\A}{\A + \C}}, {\mu := \frac{\B}{\B + \C}}$. It holds 
			\begin{align*}
			0 &< \A,\B,\C < 1 \\
			\Rightarrow 0 &< \lambda, \mu < 1,\\
			p_{A \cup C} &= \frac{\A \a + \C \c}{\A + \C} \\&= \lambda \a + (1-\lambda)\c, \\
			p_{B \cup C} &= \frac{\B \b + \C \c}{\B + \C} \\&= \mu \b + (1-\mu)\c.
			\end{align*}
			Solving the last two equations for $\c$ it holds
			\begin{align*}
			\frac{\a - \lambda \b}{1 - \lambda} &= \c = \frac{\b - \mu \a}{1-\mu} \\
			\Leftrightarrow \a-\lambda\b-\mu\a+\lambda\mu\b &= \b-\lambda\b-\mu\a+\lambda\mu\a \\
			\Leftrightarrow (1-\lambda\mu)\a &= (1-\lambda\mu)\b \\
			\overset{\mu\lambda \neq 1}{\Leftrightarrow} \a = \b \\
			\Rightarrow \c = \frac{\b - \lambda \b}{1 - \lambda} &= \b
			\end{align*}
			as stated.
		}
	\end{proof}
\end{lemma__}

\begin{definition_}
	We say that a pair of $P_T$-non-null sets $A,B \in \B$ are \textdef{alternating sets} iff $p_A \neq p_B$. 
	If alternating sets exist, then we say that $p_t$ has \textdef{model drift}.
\end{definition_}

\begin{corollary__}
	\label{thm:GeneralChangepoint}
	Let $(p_t,P_T)$ be a drift process and let $A,B \in \B$ be disjoint, alternating sets. Then $A,A^C$ or $B,B^C$ are alternating, too.
	\begin{proof}
		%
		Let $C = (A \cup B)^C$. If $C$ is a null set, then $p_A \neq p_B = p_{B \cup C}$ and hence we have that $A,A^C$ are alternating. If $C$ is not a null set, then $p_A \neq p_{B \cup C}$ or $p_B \neq p_{A \cup C}$, by Lemma~\ref{thm:ModelCoveringLemma}, and hence $A,A^C$ resp. $B,B^C$ are alternating.
	\end{proof}
\end{corollary__}

\begin{corollary__}
	\label{thm:ModelCrossCoveringCorollary}
	Let $(p_t,P_T)$ be a drift process and let $A,B \in \B$ such that $P_T(A),P_T(B) \in (0,1)$. If 
	\begin{align*}
	p_A = p_{A^C} \qquad \text{and} \qquad p_B = p_{B^C}
	\end{align*}
	then it holds $p_A = p_B$.
	\begin{proof}
		We may find pair wise disjoint sets ${D,E,F \in \B}$, such that ${A,A^C,B,B^C \in \sigma(\{D,E,F\})}$. 
		Without loss of generality we may assume ${A = D \cup E}, {A^C=F}, {B=D},{B^C=E \cup F}$. If $E$ is a null set, then trivially it holds $p_A = p_B$ otherwise we may apply Lemma~\ref{thm:ModelCoveringLemma} to see that $p_D = p_E$ and hence $p_A = p_B$.
	\end{proof}   
\end{corollary__}

We will now give a proof of Theorem~\ref{ModelDriftAndProperDriftAreEquivalent_}:
\begin{theorem_}
	\label{ModelDriftAndProperDriftAreEquivalent_}
	Let $(p_t,P_T)$ be a drift process and let $\B_0 \subset \B$ a generating set (i.e.\ $\sigma(\B_0) = \B$), which is stable under finite intersections. Then the following properties are equivalent:
	\begin{enumerate}
		\item $p_t$ has proper drift,
		\item $p_t$ has model drift,
		\item there exist alternating sets $A,A^C$, with $A \in \B_0$.
	\end{enumerate}
	\begin{proof}
		\textbf{"$3. \Rightarrow 2.$":} is clear. 
		\textbf{"$2. \Rightarrow 1.$":} Let $A,B$ be alternating sets. Assume that $p_t$ has no proper drift. By Lemma~\ref{KeyCharacterizationOfProperDrift} we have $p_t \otimes P_T = P_X \times P_T$ and hence $p_A = P_X = p_B$ which is a contradiction.
		
		\textbf{"$1. \Rightarrow 3.$":} 
		%
		%
		%
		Assume that for all $A \in \B_0$ with $P_T(A) \in (0,1)$ it holds $p_A = p_{A^C}$. Then it follows by Corollary~\ref{thm:ModelCrossCoveringCorollary} that $p_A = p_B$ for all $A,B \in \B_0$ with $P_T(A),P_T(B) \in (0,1)$. Since adding a null set to $A$ wount change $p_A$ we have that $p_A = p_B$ for all non-null sets $A,B \in \B_0$.
		
		Hence $P_X = p_A$ is well defined for any non-null set $A \in \B_0$. Now it holds 
		\begin{align*}
		(p_t \otimes P_T)(B \times C) &\overset{\text{def. } p_B}{=} P_T(B) p_B(C) \\&= P_T(B) P_X(C)
		\end{align*}
		for all $B \in \B_0, C \in \A$. Since $\sigma(\B_0 \otimes \A) = \sigma(\B \otimes \A)$ we have that $p_t \otimes P_T = P_X \times P_T$ which by Lemma~\ref{KeyCharacterizationOfProperDrift} implies that $p_t$ has no proper drift. This is a contradiction.
	\end{proof}
\end{theorem_}


We will now give a proof of Corollary~\ref{Changepoint_}:
\begin{corollary_}
	\label{Changepoint_}
	Let $(p_t,P_T)$ be a drift process and suppose that $\T \in \{ [a,b],\R_{\geq 0},\R\}, \B = \Borel(\T)$. If $p_t$ has proper drift, then there exists a change-point, i.e. it exists a $t_0 \in \R$ such that $p_{\{t < t_0\}} \neq p_{\{t \geq t_0\}}$.
	\begin{proof}
		Recall that $\{(-\infty,a) | a \in \Q \}$ is a generator of $\Borel(\R)$; therefore we may find a $t_0 \in \R$ such that $A = (-\infty,t_0) \cap \T, A^C$ are alternating sets (Theorem~\ref{ModelDriftAndProperDriftAreEquivalent_}).
	\end{proof}
\end{corollary_}

\subsubsection{Drift as non-stationarity of a stochastic process}

\begin{definition__}
	\label{def:StochProcess}
	Let $(\Omega,\mathcal{F},\P)$ be a probability space, $(\X,\A)$ be a measurable space and $\T$ be an index set. A \textdef{stochastic process} is a collection of $\T$-indexed, $\X$-valued random variables $(X_t)_{t \in \T}$. 
	By fixing a $\omega \in \Omega$ we obtain a map $X_\bullet (\omega) : \T \to \X, t \mapsto X_t(\omega)$; we refer to those maps as \textdef{the paths of $X_t$}.
	We say that $X_t$ has \textdef{$\P$-a.s. continuous paths} iff $t \mapsto X_t(\omega)$ is continuous for $\P$-a.s. all $\omega \in \Omega$.
\end{definition__}

\begin{definition_}
	Let $(\Omega,\mathcal{F},\P)$ be a probability space. A stochastic process $X_t : \Omega \times \T \to \X$ is \textdef{stationary} if \[\P \circ (X_{t_1},...,X_{t_n})^{-1} = \P \circ (X_{t_1+\tau},...,X_{t_n+\tau})^{-1}\] for all $t_1,...,t_n \in \T$, $\tau \in \T$ and $n \in \N$. 
\end{definition_}

We will now give a proof of Theorem~\ref{Stationary_}:
\begin{theorem_}
	\label{Stationary_}
	Let $X_t$ be a stochastic process. For every sequence $t_1,...,t_n \in \T$ we obtain a non-probabilistic drift process by setting $p^{(t_1,...,t_n)}_\tau = \P \circ (X_{t_1+\tau},...,X_{t_n+\tau})^{-1}$. 
	If $X_t$ is stationary then $(p^{(t_1,...,t_n)}_\tau,P_T)$ has no drift for all probability measures $P_T$ on $\T$, $t_1,...,t_n \in \T$ and $n \in \N$.\\
	Furthermore, if $P_T$ has no null sets, then the notion of stationarity of $X_t$ and $(p^{(t_1,...,t_n)}_\tau,P_T)$ having no drift for all $t_1,...,t_n \in \T$ and $n \in \N$, are equivalent. 
	\begin{proof}
		Using Lemma~\ref{ClassicalAndProbabilisticDrift} the problem boils down to remarking that the empty set always has measure zero.
	\end{proof}
\end{theorem_}

\begin{corollary__}
	\label{thm:ContinuousPathes}
	Let $X_t$ be a $\R^d$-valued stochastic process with $\P$-a.s. continuous paths. Let $P_T$ be a probability measure on $\T$. Suppose that Lebesgue-measure is absolutely continuous with respect to $P_T$. Then $X_t$ is stationary if and only if  $(p^{(t_1,...,t_n)}_\tau,P_T)$ has no drift for every sequence $t_1,...,t_n \in \T$ and $n \in \N$..
	\begin{proof}
		"$\Leftarrow$" follows by Theorem~\ref{Stationary_} and Theorem~\ref{ProbabilisticDriftAndConstantDriftAreEquivalent_}. Show "$\Rightarrow$": 
		Since $p^{(t_1,...,t_n)}_{\tau}$ has no drift we may find a $\tau_0$ such that $p^{(t_1,...,t_n)}_{\tau_0} = p^{(t_1,...,t_n)}_{\tau}$ for $P_T$-a.s. all $\tau \in \T$ (Corollary~\ref{ConstantDriftIsRealizedAtSomePoint}). 
		
		It remains to prove that $\tau \mapsto p^{(t_1,...,t_n)}_\tau$ is continuous with respect to total variation norm, then $t \mapsto \Vert p^{(t_1,...,t_n)}_\tau - p^{(t_1,...,t_n)}_{\tau_0} \Vert$ is continuous. Since it is equals 0 $P_T$-a.s. and continuous it follows that it is 0 everywhere, since Lebesgue-measure is absolutely continuous with respect to $P_T$, which implies $p^{(t_1,...,t_n)}_\tau = p^{(t_1,...,t_n)}_{\tau_0}$ for all $\tau \in \T$. 
		
		By the triangle inequality $\tau \mapsto (X_{t_1+\tau},...,X_{t_n+\tau})$ is $\P$-a.s. continuous, it is therefore enough to show that a stochastic process with a.s. continuous paths has continuously changing marginal distributions, but this is well known (also known as: sample continuity implies continuity in distribution).
	\end{proof}
\end{corollary__}

\subsubsection{Drift as dependency between data and time}

\begin{definition_}
	Let $(p_t,P_T)$ be a drift process and let $(X,T) \sim p_t \otimes P_T$ a pair of random variables. We say that $p_t$ has \textdef{dependency drift} if $X$ and $T$ are not independent. 
\end{definition_}

We will now give a proof of Theorem~\ref{DriftAsDependency_}:
\begin{theorem_}
	\label{DriftAsDependency_}
	Let $(p_t, P_T)$ be a drift process. Then $p_t$ has proper drift if and only if it has dependency drift.
	\begin{proof}
		Let $(\Omega,\mathcal{F},\P)$ be the underlying probability space, i.e. $X$ and $T$ are measurable maps from $\Omega$ to $\X$ resp. $\T$. $X$ and $T$ are independent if and only if \[(p_t \otimes P_T)(A \times B) = \P_{X,T}(A \times B) = \P_X(A)\P_T(B)\] holds for all $A \in \A, B \in \B$. By setting $A = \X$ we obtain $\P_T = P_T$ and therefore $p_t \otimes P_T = \P_X \times P_T$ which, by Lemma~\ref{KeyCharacterizationOfProperDrift}, holds if and only if $p_t$ has no proper drift.
	\end{proof}
\end{theorem_}

\subsubsection{Linear-DriFDA}

We will now give a proof of Lemma~\ref{LinearDriFDAComputation_}:
\begin{lemma_}
	\label{LinearDriFDAComputation_}
	Let $(\Omega,\mathcal{F},\P)$ be a probability space, $\S_D$ and $\S_I$ be independent, $\R^{n_D}$- resp. $\R^{n_I}$-valued random variables. Let $A : \R^{n_D+n_I} \to \R^d$ be a linear map.
	
	Denote by $\S := (\S_D,\S_I)$, $X := A\S$, $X_D = \E[X | \S_D]$ and $X_I = \E[X | \S_I]$. Then it holds
	\begin{align*}
	X_D &= A \S_D + A\E[\S_I],\\X_I &= A \S_I + A\E[\S_D],\\X_D + X_I &= X + \E[X].
	\end{align*} 
	\begin{proof}
		Without loss of generality we may assume that $\S = \S_D + \S_I$, i.e. $\S_D$ and $\S_I$ "use different dimension". Now its a computation
		\begin{align*}
		X_D &= \E[A \S | \S_D] \\&= A(\E[\S_D| \S_D] + \E[\S_I | \S_D]) \\&= A\S_D + A\E[\S_I] 
		\\X_I &= \cdots = A\S_I + A\E[S_D]
		\\\Rightarrow X &= A\S \\&= A\S_D + A\S_I \\&= (X_D - A\E[\S_I]) + (X_I - A\E[\S_D]) \\&= X_D+X_I - \E[A\S] \\&= X_D+X_I - \E[X].
		\end{align*}
		Note that we dropped the dimensions containing $T$ for simplicity. 
	\end{proof}
\end{lemma_}

\newpage

\subsection{Algorithms}
\begin{algorithm}
	\caption{$k$-curve-DriFDA}
	\label{kCDriFDAAlg}
	\newcommand{\D}{\mathcal{D}}
	\begin{algorithmic}[1]
		\Procedure{$k$-curve-DriFDA: Drifting Feature Decomposition Analysis via $k$-curves}{$(x_j,t_j)$ data steam, $k$ number of curves} \;
		\State $\D \gets \emptyset$ \;
		\State Initialize $\mu_i, i=1...,k$ using $k$-means. \;
		\While{Not at end of stream} \;
		\State Draw next batch $\D_\text{new}$ from stream \;
		\State $\D \gets \D \cup \D_\text{new}$ \;
		\While{$\mu_i$ not converged} \;
		\For{$i = 1,...,k$} \State $\D_i \gets \emptyset$ \EndFor
		\For{$(x,t) \in \D$} \;
		\State $i^* \gets \argmin_i \Vert x - \mu_i(t) \Vert$ \;
		\State $\D_{i^*} \gets \D_{i^*} \cup \{(x,t)\}$ \;
		\EndFor
		\For{$i = 1,...,k$} \State Retrain $\mu_i$ using $\D_i$ \EndFor
		\EndWhile
		\EndWhile
		\State $\textbf{return } (\mu_i)_{i=1,...,k}$ \;
		\EndProcedure
	\end{algorithmic}
\end{algorithm}

\subsection{Visualization of DriFDA}

\begin{figure}[!h]
	\centering
	\includegraphics[width=0.5\textwidth]{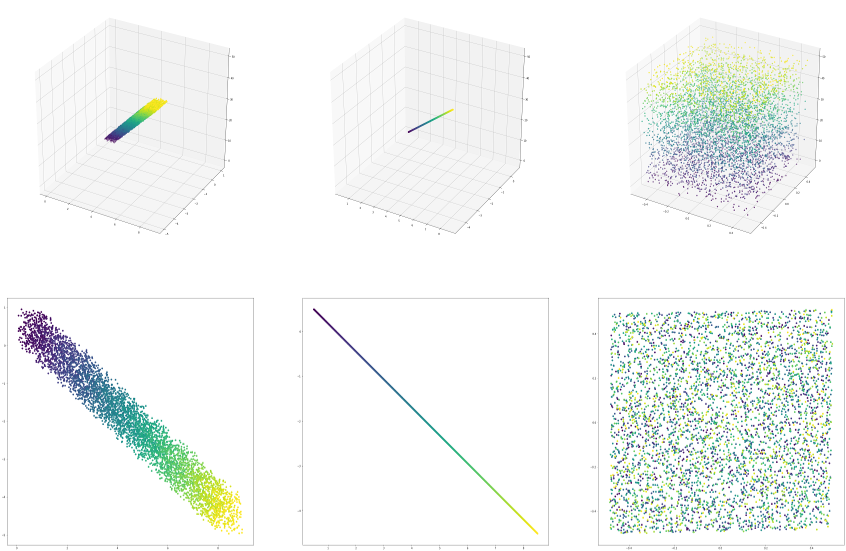}
	\caption{Linear-DriFDA applied to square data set}
\end{figure}

\begin{figure}[!h]
	\centering
	\includegraphics[width=0.5\textwidth]{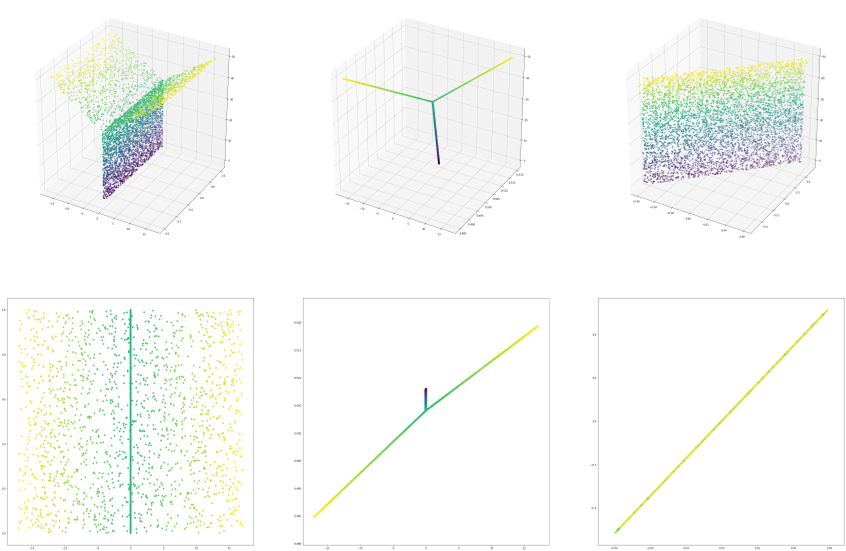}
	\caption{Linear-DriFDA applied to Y data set}
\end{figure}

\end{document}